\documentclass[letterpaper, 10pt, conference]{ieeeconf}  % Comment this line out if you need a4paper

\IEEEoverridecommandlockouts                              % This command is only needed if 
\usepackage{amsmath} % assumes amsmath package installed
\usepackage{amssymb}  % assumes amsmath package installed
\usepackage[pdftex]{graphicx}
\usepackage{breqn}
\usepackage{algorithm}
\usepackage{algpseudocode}
\usepackage{hyperref}
\usepackage{balance}
\usepackage{soul} %for strikethrough
\newcommand{\mc}[1]{{\mathcal{#1}}}

\newcommand{\COMMENT}[1]{}

\usepackage{pgfplots}
\pgfplotsset{compat=newest}
\definecolor{mLightGreen}{HTML}{14B03D}
\definecolor{mDarkRed}{HTML}{a2282f}
\tikzstyle{mydashdot}=[dash pattern=on 6pt off 2pt on \the\pgflinewidth off 2pt]
\usetikzlibrary{matrix}
\usepgfplotslibrary{groupplots}
\pgfplotsset{every tick label/.append style={font=\tiny}}
\pgfplotsset{ylabsh/.style={every axis y label/.style={at={(0,0.5)}, xshift=#1, rotate=90}}}  
\usepackage{setspace}
\newtheorem{definition}{Definition}

\usepackage{xcolor}
\newcommand{\todo}[1]{{\textcolor{red}{#1}}}
\newcommand{\edit}[1]{{\textcolor{black}{#1}}}
\newcommand{\revision}[1]{{\textcolor{black}{#1}}}
\graphicspath{{figures/}}

\hypersetup{
	colorlinks,
 	linkcolor={black},
 	citecolor={black},
 	urlcolor={black}
}

\usepackage{subfiles}

\usepackage{fancyhdr}

\fancypagestyle{firstpage}{%
  \lhead{This work has been submitted to the IEEE for possible publication. Copyright may be transferred without notice, after
which this version may no longer be accessible}
}
\title{\vspace{0.3cm}Multi-Robot Coordination and Cooperation with Task Precedence Relationships\vspace{-0.25cm}}
\author{Walker Gosrich$^{1}$, Siddharth Mayya$^{2}$, Saaketh Narayan$^{1}$, Matthew Malencia$^{1}$,\\
Saurav Agarwal$^{1}$, and Vijay Kumar$^{1}$% <-this % stops a space
\thanks{We gratefully acknowledge the support of ARL DCIST CRA W911NF-17-2-0181. This material is based upon work supported by the National Science Foundation Graduate Research Fellowship.}%
\thanks{$^{1}$W. Gosrich, S. Narayan, M. Malencia, S. Agarwal, and V. Kumar are with the GRASP Laboratory, University of Pennsylvania, Philadelphia, PA, USA {\{\href{mailto:gosrich@seas.upenn.edu}{gosrich}, \href{mailto:saaketh@seas.upenn.edu}{saaketh}, 
\href{mailto:malencia@seas.upenn.edu}{malencia}, \href{mailto:sauravag@seas.upenn.edu}{sauravag},  \href{mailto:kumar@seas.upenn.edu}{kumar}\}@seas.upenn.edu }}%
\thanks{$^{2}$S. Mayya is with Amazon Robotics, North Reading, MA, USA (\href{mailto:mayya.siddharth@gmail.com}{mayya.siddharth@gmail.com}). This work is not related to Amazon.}%
}

\begin{document}

\maketitle
\thispagestyle{firstpage}
\pagestyle{empty} 

%%%%%%%%%%%%%%%%%%%%%%%%%%%%%%%%%%%%%%%%%%%%%%%%%%%%%%%%%%%%%%%%%%%%%%%%%%%%%%%%
\begin{abstract}
%%%%%%%%%%%%%%%%%%%%%%%%%%%%%%%%%%%
%%% Developed from Vj's version %%%
%%%%%%%%%%%%%%%%%%%%%%%%%%%%%%%%%%%

We propose a new formulation for the multi-robot task planning and allocation problem that incorporates
(a)~precedence relationships between tasks;
(b)~coordination for tasks allowing multiple robots to achieve increased efficiency; and
(c)~cooperation through the formation of robot coalitions for tasks that cannot be performed by individual robots alone.
In our formulation, the tasks and the relationships between the tasks are specified by a task graph.
We define a set of reward functions over the task graph's nodes and edges. These functions model the effect of robot coalition size on task performance while incorporating the influence of one task's performance on a dependent task.
%We define a set of reward functions over the task graph's nodes and edges that model the influence of one task's performance on the next, and the effect of forming coalitions on task performance.
Solving this problem optimally is NP-hard.
However, using the task graph formulation allows us to leverage min-cost network flow approaches to obtain approximate solutions efficiently.
Additionally, we explore a mixed integer programming approach, which gives optimal solutions for small instances of the problem but is computationally expensive. We also develop a greedy heuristic algorithm as a baseline.
Our modeling and solution approaches result in task plans that leverage task precedence relationships and robot coordination and cooperation to achieve high mission performance, even in large missions with many agents.

\end{abstract}

\section{Introduction}
\label{sec:intro}
\begin{figure}
	\centering 
	\includegraphics[trim={0.0cm 0.0cm 0.0cm 0.0cm},clip,width=\linewidth]{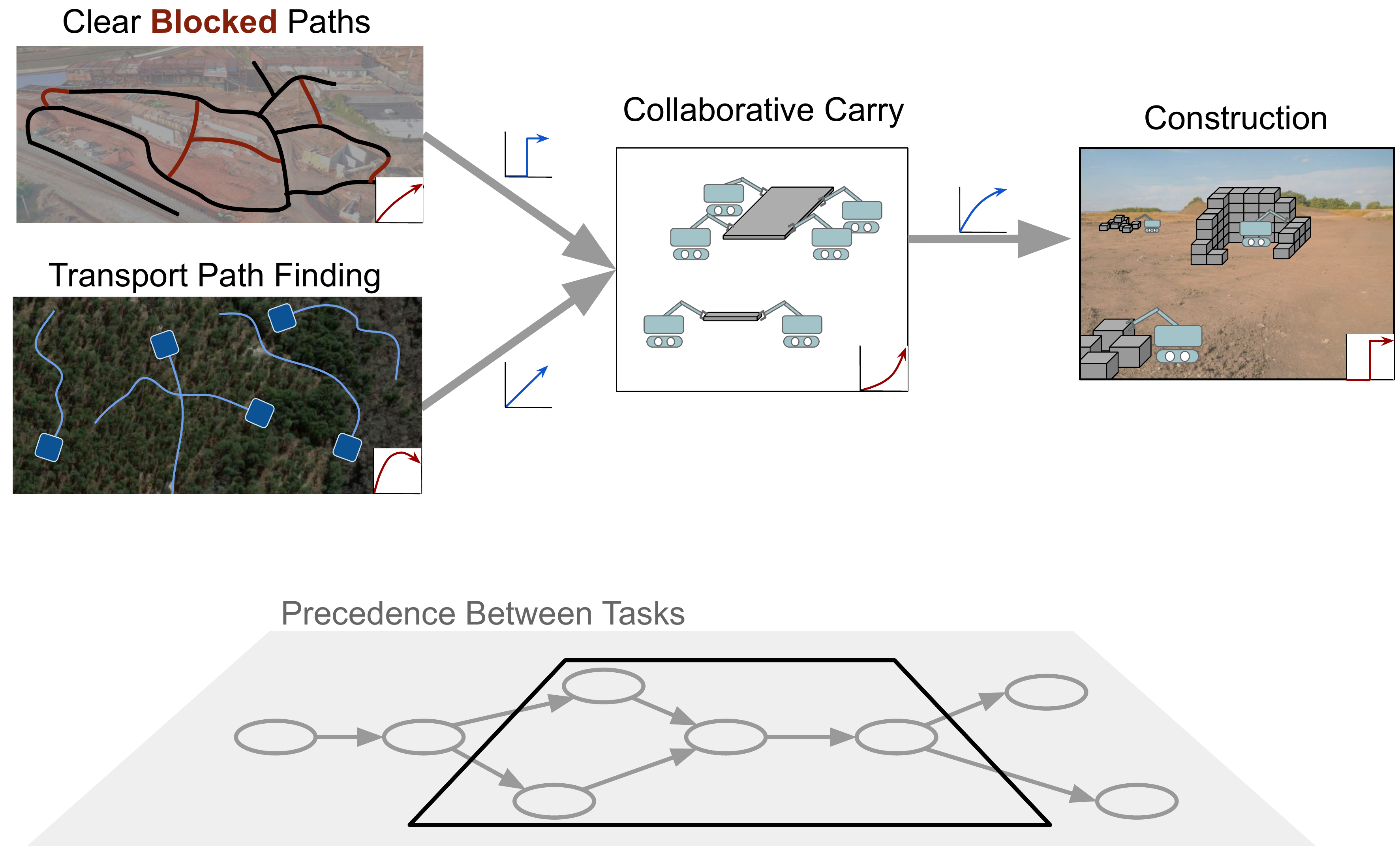}	\caption{\label{fig:intro_illustration} The task graph (\textit{bottom}) shows precedence relationships between tasks in a mission. Highlighted is a subset of the tasks, showing coalition functions (\textit{red}) and precedence relationships (\textit{blue}). For example, before transporting construction materials, robots may need to find or clear paths through a debris-filled construction site. The quality of these paths directly impact the team's ability to transport material. Additionally, the performance of transporting material, e.g., whether any materials are damaged, impacts how quickly and how well the team performs a following construction task.}
\end{figure}

Multi-robot systems offer versatility to accomplish large missions by forming teams to efficiently complete tasks. A fundamental challenge in multi-robot missions is to allocate tasks to robots, referred to as the multi-robot task allocation (MRTA) problem.
This paper considers a generalization of the MRTA problem---tasks have precedence relationships among them, i.e., the quality of a task's execution may depend on how well a related task is executed.
For example,  in an autonomous construction mission, an area must be cleared and leveled, and materials gathered before construction takes place, as illustrated in Figure~\ref{fig:intro_illustration}.
In such a scenario, the quality with which prior tasks are completed impacts the ability of the robot team to complete future tasks, e.g., a well-leveled foundation will result in a better overall performance of the construction mission.
Thus, it is imperative to model relationships among tasks, especially the {\em influence} of a preceding task on subsequent tasks.
%The paper addresses such relationships among tasks by modeling them as a task graph and introduces an influence function to quantify the effect of a preceding task on future tasks.

%There are two critical components in a multi-robot mission: {\em coordination}---the distribution of robots among tasks, and {\em cooperation}---the coalition of robots into subteams to perform tasks.
The other important generalization of MRTA we consider is that a coalition of multiple robots can work on a task together. On some tasks, such as multi-robot coverage, robots may {\em coordinate} to achieve improved results. Other tasks, such as collaborative carry, require multiple robots to {\em cooperate} to perform the task when a single robot cannot (e.g., Figure~\ref{fig:intro_illustration}). 
%\edit{There are two critical components in a multi-robot mission: {\em coordination} for tasks allowing multiple robots to achieve increased efficiency in proportion to the number of the robots; and {\em cooperation} through the formation of robot coalitions for tasks that cannot be performed by individuals but can be completed by a group.}
The standard MRTA problem considers only the coordination of robots~\cite{gerkey2004formal}.
In the simplest case of single robot teams, this problem can be posed as an assignment problem~\cite{BurkardDM12}.
\revision{However, some tasks require multi-agent teams to be effectively executed. Additionally, multi-agent teams can bring greater efficacy to tasks such as sensor coverage, exploration, and resource gathering when these tasks are constrained by a time budget.}
Such scenarios raise two challenges:
(1)~the estimation of task utility by modeling the relationship between robot coalitions and tasks, and
(2)~the assignment of robot coalitions to tasks under temporal and resource constraints.
A desirable characteristic of a task planner is to be able to reason about resource allocation, e.g., if task relationships indicate that there is a mission-critical task, a larger robot coalition may be allocated to that task, even at the expense of other tasks' performance. 
Hence, there is a need for a formulation that models the relationship of robot coalitions to tasks and unifies coordination and cooperation of robots for \revision{MRTA}.
%This paper presents a novel formulation that unifies the coordination and cooperation of robots for the MRTA problem.

%\todo{\st{These challenging aspects of the task allocation problem have been considered separately in the literature but seldom together.}}
%The coalition formation community offers many approaches to the problem when only an instantaneous assignment (rather than a full task allocation schedule) is required, or when inter-task relationships are not present.
%Other research considers task allocation with precedence constraints, but often treats the problem as constraint satisfaction without measuring solution quality \cite{deng2019compiler}, considers precedence constraints as binary, or considers task rewards as binary.
This paper presents a novel formulation that unifies precedence relationships among tasks, coordination and cooperation among robots through coalitions, and allocation of coalitions to tasks.
The formulation is applicable to complex scenarios relevant to autonomous construction, precision agriculture, and industrial robotic applications.
The contributions are:
\begin{enumerate}
    \item A modular and expressive mission model defined over a {\em task graph} and a reward function (Section~\ref{sec:mission_model}).
   The task graph captures the precedence relationships among tasks.
   The reward function quantifies the utilities of the tasks and is characterized by {\em influence functions}---modeling of relationships among tasks, {\em aggregation functions}---effect of all preceding tasks on a dependent task, and {\em coalition functions}---modeling of task execution efficacies with robot coalition size.
   \item A formulation of the problem as a min-cost network flow by leveraging the task graph (Section~\ref{sec:flow-solution}).
    This enables the allocation of robots to sequences of connected tasks in the task graph, and scales well with the number of tasks and robots in the mission.
    %The approach scales well with the number of tasks and is agnostic to the number of robots.
    \item We benchmark our flow-based solution approach with a mixed integer approach and a greedy heuristic algorithm (Section~\ref{sec:experiments}).
    The comparison is performed on over 300 randomly generated missions, including missions drawn from an autonomous construction simulator.
    Our flow-based approaches significantly outperform the mixed integer benchmark in large problem sizes, and perform similarly in the smaller problem instances where the mixed integer approach can compute a non-trivial solution. Our flow-based non-linear programming (NLP) approach significantly outperforms the greedy approach across the problem domain.
\end{enumerate}

\section{Previous Work}
\label{sec:background}
This work focuses on multi-task missions composed of multi-robot tasks with inter-task precedence relationships.
\edit{In applications where these mission characteristics are common, such as autonomous construction and assembly \cite{knepper2013ikeabot}, agriculture \cite{mao2021research}, and other complex multi-robot missions, efficient allocation of robot coalitions that enable \textit{coordination} and \textit{cooperation} is essential.}
%In applications where these mission characteristics are common, such as autonomous construction and assembly, agriculture, and other complex multi-robot missions, robot \textit{coordination} is required to handle temporal relationships between tasks, and robot \textit{cooperation} is required to form appropriate coalitions for multi-robot tasks.
In prior work, each of these characteristics has been considered extensively but rarely together in a cohesive model.

Many approaches allocate robots among interrelated tasks without considering multi-robot coalitions. Some approaches formulate the problem purely as a constraint satisfaction problem (CSP) \cite{deng2019compiler, dogar2019multi}, while others hybridize Mixed Integer Linear Programming (MILP) with CSP in order to reason about task utility \cite{gombolay2013fast}. In operations research \cite{NUNES2017}, the task allocation problem with inter-task relationships is most similar to the Vehicle Routing Problem with Time Windows~\cite{kolen1987vehicle}, with additional ordering constraints \cite{bredstrom2008combined}, \edit{though these formulations do not consider multi-agent tasks or expressive inter-task relationships.}

%Many approaches coordinate robots among interrelated tasks without considering multi-robot coalitions. Some approaches formulate the problem as a constraint satisfaction problem (CSP) \cite{deng2019compiler, dogar2019multi}, in which task utility is not considered. Others hybridize Mixed Integer Linear Programming (MILP) with CSP in order to reason about task utility \cite{gombolay2013fast}. In operations research \cite{NUNES2017}, the task allocation problem with inter-task relationships is most similar to the Vehicle Routing Problem with Time Windows (VRPTW) \cite{kolen1987vehicle}, with additional ordering constraints \cite{bredstrom2008combined}. This optimization perspective offers valuable tools for simultaneous scheduling and allocation, but has several distinctions from robotics applications: formulations typically assume an unlimited number of agents available, do not consider coalition formation for multi-agent tasks, and have a binary representation of task relationship. 

In the computer science literature, Multi-Agent Planning (MAP) with \textit{joint actions} expresses the tight coupling between tasks that we represent with precedence relationships. Several approaches (\cite{brafman2014distributed, shekhar2020signaling}) consider this combination of challenges but consider coalitions in a limited fashion. Some approaches \revision{(\cite{TERESHCHUK2021102154, smith2019real, ponda2010decentralized}) % todo add other citations if possible
consider soft precedence constraints encoded via time-varying task rewards or imposing costs. Other methods \cite{smith2019real, wang2022heterogeneous} model inter-task relationships among single-agent tasks in detail on a heterogeneous graph, with nodes representing tasks, agents, and locations.}

\revision{Other approaches focus on \textit{cooperation}, modeling the relationship between the coalition assigned to a task and task performance.} Some works model reward as a function of the number of homogeneous robots assigned to complete the task \cite{Korsah2013, seenu2020review}. Others consider task success based on required attributes aggregated among a team of robots with heterogeneous capabilities \cite{prorok2017impact}. Some such methods perform \revision{short horizon} task assignment \cite{mayya2021resilient}, which allows robust response to dynamic missions but does not permit reasoning about interrelated tasks over a time horizon. Others do not consider task reward explicitly~\cite{messing2022grstaps}. 

Some approaches, like ours, consider the challenging intersection of these problem characteristics, requiring a system to reason about intersecting robot schedules and fluidly form coalitions among different tasks over an extended time horizon.
% REMOVED 
%In the taxonomy developed by Nunes et al.~\cite{NUNES2017}, this problem is identified as MT-MR-TA[XD]: task allocation with multi-task robots (MT), multi-robot tasks (MR), a time-extended assignment of robots (TA) beyond the immediate available tasks, and cross-schedule dependencies (XD) between tasks.
\revision{The approaches that consider both task order and coalition typically use binary task rewards \cite{capezzuto2020anytime}, and some additionally cannot scale to problems larger than a few robots and tasks \cite{ramchurn2010coalition, korsah2012xbots}.} In this work, we present an expressive model for task reward that expands these models, and we present solution methods that scale to large problem sizes.

% \begin{figure}
% 	\centering 
% 	\includegraphics[trim={0.0cm 0.0cm 0.0cm 0.0cm},clip,width=\linewidth]{figures/ag_example.png}
% 	\caption{\label{fig:task_graph} [PLACEHOLDER: figure that demonstrates a task graph laid out over a real-world example problem. Ideally the test case we choose for communicating this work]}
% \end{figure}

%\todo{WG: Flow stuff, min-cost flow.}~\cite{Orlin93}

\section{Modeling Coalition Performance and Task Inter-dependencies}
\label{sec:mission_model}
\newcommand*{\Mbar}{\overline{\mc {M}}}
\newcommand*{\Njin}{\mc N_j^{\mathrm{in}}}
\newcommand*{\Njout}{\mc N_j^{\mathrm{out}}}

In this section, we introduce \emph{task graphs}, which encode the structure of task relationships, and we define a {\em task reward model} over the task graph. \revision{Taken together, these models can represent complex missions for homogeneous robot teams.} 

Let $\mc{T} :=\{\mc{T}_1,\ldots,\mc{T}_M\}$ represent the set of $M$ available tasks in the environment, and $\mc{M} :=\{1,\ldots,M\}$ represent the index set of tasks.
Each task $\mc T_j$ has a constant duration~$d_j$ and yields a reward $r_j$ upon completion.
\edit{The reward is a function of the assigned robot coalition and the rewards of related preceding tasks, as defined later in the section.} \revision{Note that a task does not start until the complete coalition assigned to the task has formed.}
%The reward is a function of prior task performance and the coalition assigned to the task, as is defined later in the section. 

\begin{definition}[Task Graph]
Given a set of tasks $\mc{T}$, the task graph is a directed acyclic graph $\mc{G}_T = (\Mbar, \mc{E})$, where $\Mbar = \{0,1,\ldots,M\}$ is the vertex set corresponding to tasks and $\mc{E}$ is the edge set corresponding to task relationships.
The vertex $0$ represents a virtual ``source'' node, and all other vertices correspond to actual tasks.
We add a directed edge $(i, j)$ to the set $\mc E$ if there exists a precedence relationship between tasks $\mc T_i$ and $\mc T_j$ such that the reward $r_i$ associated with task $\mc T_i$ impacts the reward $r_j$ from task $\mc T_j$. 
We add an edge from the source node $\mc{T}_0$ to all tasks without preceding neighbors, resulting in a directed acyclic graph.
\end{definition}%
%\begin{definition}[Task Graph]
%Given a set of tasks $\mc{T}$, let $\mc{G}_T = (\Mbar, \mc{E})$ be a directed graph, where $\Mbar = \{0,1,\ldots,M\}$ is the vertex set and $\mc{E}$ is the corresponding edge set.
%Vertices $1,\ldots,M$ represent the actual tasks, and vertex $0$ represents a virtual ``source" node.
%If a precedence relationship exists such that the reward $r_i$ from task $\mc{T}_i$ impacts task $\mc{T}_j$ (e.g., a passage must be cleared before before entering and surveying a room), we represent this relationship via a directed edge $(i, j) \in \mc{E}$ from task $\mc{T}_i$ to task $\mc{T}_j$.
%We connect source node $\mc{T}_0$ to all tasks without preceding neighbors. \edit{This results in a Directed Acyclic Graph (DAG).}
%\end{definition}

Furthermore, let $\mc R$ represent the index set of $N$ robots available to execute the tasks. Let $x_k^r \in \{0, 1\}$ indicate whether a robot $r\in \mc R$ executes task~$k\in \overline{\mathcal M}$ during the mission. We define a robot coalition $\mc{C}_j$ as the set of robots allocated to task $\mc{T}_j$. Furthermore, let $C_j = |\mc{C}_j|$ represent the size of the coalition, where $|\cdot|$ is the set cardinality operator.

\subsection{Task Reward Model}
\label{sec:task_model}
We present a model for the reward $r_j$, associated with the task $\mc T_j\in \mc T$.
The task reward model comprises three functions:
(1)~a {\em coalition function} $\rho_j$,
an (2)~{\em aggregation function} $\alpha_j$, and
(3)~{\em influence functions} $\delta_{ij}$.

%We present here a model for the reward $r_j$ of the tasks $\mc{T}_j \in \mc{T}$.
%This model represents the reward as a function of the coalition of robots $\mc{C}_j$, corresponding to the number of agents that complete a task $\mc{T}_j$, and of the reward accrued by prior tasks related by precedence relationships.  

%\begin{enumerate}
%    \item \todo{To} represent the contribution of each task $j$ towards the overall mission success with its reward $r_j$, such that the overall mission success is represented by the sum of all task rewards $$\sum_{j\in \mathcal{M}}r_j$$
    
%    \todo{Modeling the individual contribution of each task towards the overall mission represents tasks that both contribute to future task success, and that are independently important. One example is a coverage task that is later exploited for resource collection or for route planning. The data collected in the coverage task could be independently important in the mission, while facilitating success on the routing or collection tasks. }
%    \item \todo{To} represent the reward $r_j$ as a function of two variables: the total robot coalition assigned to task $j$, $f_{\rightarrow j}$, and the rewards earned by all tasks constrained by precedence constraints to precede $j$; that is, all incoming neighbors into node $j$ in the graph $G_T$
%\end{enumerate}

\begin{definition}[Task Coalition Function]
% The task coalition function $\rho_j: \mathbb{R} \rightarrow \mathbb{R}$ takes as input the coalition $\mc{C}_j$ assigned to task $\mc{T}_j$, and has an output that represents the effectiveness of the robot coalition at accomplishing the given task.
Given a robot coalition $\mc{C}_j$ assigned to task $\mc{T}_j$, the task coalition function $\rho_j(C_j): \mathbb{R} \mapsto \mathbb{R}$ returns a scalar that represents the effectiveness of the robot coalition at accomplishing the given task.
\end{definition}

\begin{figure}[tbp]
\vspace{0.1cm}
	\centering 
	\includegraphics[trim={0.0cm 0.0cm 0.0cm 0.0cm},clip,width=\linewidth]{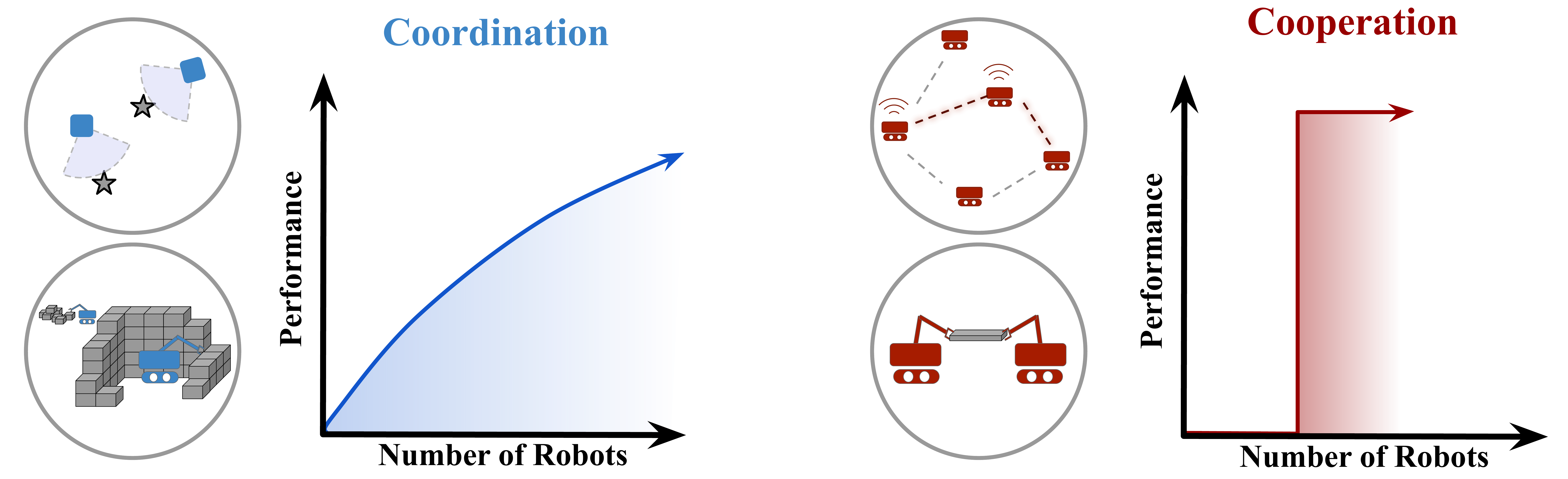}
	\caption{\label{fig:coalition} Figure adapted from \cite{prorok2021beyond}. Two types of coalition function $\rho$. \textit{Left} tasks such as coverage control and construction have a subadditive coalition function, as larger coalitions improve performance. \textit{Right} network connectivity tasks and cooperative transport are represented by a ``step'' coalition function. When a critical mass of robots is collected in the coalition, the task can successfully be performed.}
% 	network becomes connected and maximum reward is accrued.}
\end{figure}

\begin{figure}[tbp]
	\centering 
	\includegraphics[trim={0.0cm 1.7cm 0.0cm 0.55cm},clip,width=\linewidth]{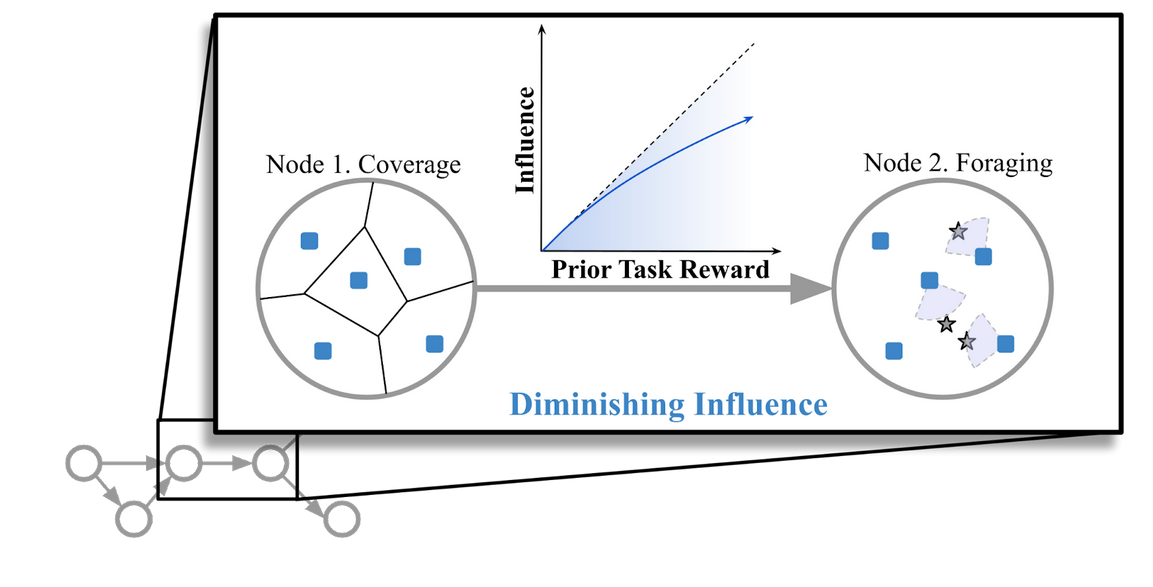}
	\caption{\label{fig:inluence} The influence function, illustrated here, models the relationship between the reward at the preceding (or \textit{influencing}) task and the reward at the current task. A better result on a coverage task would result in higher performance in the foraging task (exploiting the information accrued during coverage). This is modeled by a subadditive influence function.} \vspace{-0.5cm}
\end{figure}

%\todo{Description of the value of modeling modular task rewards is omitted here.}
% \todo{We recognize that many other factors, including environmental factors, individual robot effectiveness, etc. may impact the reward accrued by tasks. However, we neglect these terms in this model, because such factors are often unknown at planning time.}

% The coalition function provides an expressive model for the relationship between robot coalition and task reward. It can model performance that is both linear and non-linear, e.g., Figure \ref{fig:coalition} shows tasks with diminishing returns in the number of robots and tasks that require a threshold number of robots to complete \cite{dutta2019one}.

% Prior work implicitly assumes a linear coalition function, where the reward or performance of a task increases linearly with the size of its assigned coalition \cite{dutta2019one,zitouni2019fa}. Instead, our task coalition functions can model more complex coalition-reward relationships that are common to robotic tasks, as seen in collaborative tasks with diminishing returns (see Figure \ref{fig:coalition} left) or cooperative tasks that can only be achieved by a threshold number of robots (see Figure \ref{fig:coalition} right).

The coalition function provides an expressive model for the relationship between the robot coalitions assigned to a task and the resulting task reward. While prior works \cite{dutta2019one,zitouni2019fa} implicitly assume that the coalition functions are linear, \revision{ this paper uses task coalition functions to model both linear and non-linear performance characteristics.} Examples of these more complex coalition-reward relationships are seen in Figure~\ref{fig:coalition} with common robotic tasks such as collaborative tasks with subadditive reward (left) and cooperative tasks that can only be achieved by a threshold number of robots (right).
%\todo{[Use subadditive instead of diminishing returns?]}

We now model the impact of the precedence relationships on the reward $r_j$ for each task $\mc{T}_j$. 
\begin{definition}[Task Influence Function]
The influence function $\delta_{ij} (r_i): \mathbb{R} \mapsto \mathbb{R},\; (i,j)\in\mathcal{E}$ returns a scalar that quantifies the extent to which the performance of the robots on task $\mc T_i$ hinders or facilitates the execution of task $\mc T_j$. In other words, it models the influence of task reward $r_i$ on $r_j$.
\end{definition} \par 
% The task influence function models how the performance of a preceding task can hinder or facilitate success in a following task.
The task influence function provides a generalization of precedence constraints commonly studied in task allocation models. Typical approaches model task precedence as a binary construct: a dependent task may be executed only if a preceding task is completed with performance above a certain threshold. In contrast, representing $\delta_{ij}$ as a function enables the modeling of complex relationships between tasks often found in real-world scenarios. For instance, as shown in Figure~\ref{fig:inluence}, the performance of an exploration or a coverage task could be sub-linearly related to the performance of a subsequent transportation or foraging task. Similarly, the performance of a foundation leveling task could influence the performance of future construction tasks on top of that foundation. Furthermore, our formulation can also represent a classical ``binary'' precedence constraint by choosing $\delta_{ij}$ as a step function. \par 

When a task $\mc{T}_j$ has multiple incident edges, the outputs of these multiple influence functions must be aggregated over the set of incoming precedence neighbors $\Njin= \{\mc{T}_i\mid (i,j) \in \mc{E}\}$. 
\begin{definition}[Task Influence Aggregation Function]
The task influence aggregation function $\alpha_j: \mathbb{R}^{\lvert\Njin\rvert} \mapsto \mathbb{R}$ takes as an input the set of influencing rewards $ \{r_i \mid i \in \Njin\}$ and outputs a value representing the total influence of preceding tasks on the reward $r_j$ associated with task $\mc T_j$. 
\end{definition}

For instance, the task influence aggregation function can be set to $\sum_{i \in \Njin}\delta_{ij}(r_i)$ to represent \revision{the case where completing any preceding task can result in reward on task $\mc{T}_j$}, or to $\prod_{i \in \Njin}\delta_{ij}(r_i)$ to represent \revision{the case when all preceding tasks must be sufficiently completed to achieve favorable performance on task $\mc{T}_j$}.

\subsection{Composing Final Task Rewards} 
The reward $r_j$ accrued by each task $\mc{T}_j$ represents the quality of the outcome of the task. 
%$r_j$ is a function of the coalition function output $\rho_j(C_j)$ and the aggregation of the influence functions $\delta_ij(r_i)$ from all tasks $i$ in $\mc{N}_j^{in}$. 

\begin{equation} \label{eq:reward_dynamics}
    r_j = \rho_j(C_j) \; \ddag_j \; \alpha_j\left(\{\delta_{ij}(r_i)\mid i \in \Njin\}\right)
\end{equation}
In the context of task $\mc{T}_j$, the symbol $\ddag_j$ represents a function that combines outputs of the coalition function $\rho_j$ and aggregated influence function $\alpha_j$. For example, summing $\rho_j$ and $\alpha_j$ can represent a ``soft'' constraint, where the reward is non-zero even if precedence-related tasks are incomplete.
Alternatively, the product of $\rho_j$ and $\alpha_j$  could be interpreted as a ``hard'' constraint, where a zero aggregated influence value results in a task reward of zero.
%A choice of \textbf{multiplication} as the $\ddag$ operation represents a ``hard'' constraint, where an aggregated influence output of zero yields zero reward. \todo{cite} A choice of a  \textbf{sum} operation represents a ``soft'' constraint, where $r_j$ may be non-zero even if the aggregated influence output is zero.

Taken together, the task graph $\mc{G}_T$, the set of coalition, aggregation, and $\ddag$ functions on its vertices, $\{\rho_j, \alpha_j, \ddag_j \}, \forall j\in\mc{M}$, and the set of influence functions defined over its edges $\{\delta_{jk}\}, \forall (j, k)\in\mc{E}$ completely specify the mission.

\subsection{Problem Statement, Characteristics, and Assumptions} \label{subsec:ps}
The primary objective of the MRTA problem is to generate a set of feasible robot-task assignments $x^r_k, \forall r\in\mc{R}, k\in\mc{M}$, forming a set of coalitions $\mc{C}$ that maximizes the rewards obtained over all tasks: 
\begin{equation}\label{eq:cost}
    \max_{\mc{C}_j, \forall j\in \mc{M}} \sum_{j=1}^{M} r_j
\end{equation}
while satisfying the following constraints:
\textit{(i)} the last task is completed by the robots before the makespan constraint time $T_f$,
\textit{(ii)} robots may only work on one task at a time,
\textit{(iii)} the robot-task assignments respect the precedence relationships defined over the graph edges $(i,j) \in \mc{E}$,
\textit{(iv)} the task coalitions do not exceed the total number of robots~$N$.
 \revision{We assume that all reward functions are non-negative.}
 See Sections~\ref{sec:flow-solution} and~\ref{sec:minlp} for specific instantiations of these constraints.
%\todo{How is makespan calculated? Equation formalizing the makespan constraint}

% This problem is NP-hard, which can be shown by reducing an instance of the problem to an instance of a known NP-had problem~\cite{Karp1972}.
%The above problem is NP-hard.
%Similar problems with simpler reward models have been shown to be a generalization of the team orienteering problem \cite{ramchurn2010coalition}, or the traveling salesperson problem with time windows \cite{NUNES2017}, both of which are NP-hard problems.
% An instance of our problem reduces to the NP-hard 0-1 knapsack problem~\cite{gens1980complexity} by considering the \textit{single-robot} case where each task corresponds to an item, task durations are weights, and rewards are constant.
%\revision{This time-extended MRTA problem is NP-hard, which can be shown by reducing an instance of the NP-hard 0-1 knapsack problem~\cite{gens1980complexity} to an instance of this problem by considering the \textit{single-robot} case where each task corresponds to an item in the knapsack problem, task durations are item weights, task rewards are item values, and the time budget is the weight capacity of the knapsack.}
\revision{This MRTA problem is NP-hard, which can be shown by reducing an instance of the NP-hard 0-1 knapsack problem~\cite{gens1980complexity} to an instance of this problem by considering the \textit{single-robot} case where each item in the knapsack problem corresponds to a task in the MRTA problem, item weights are task durations, item values are task rewards, and the weight capacity of the knapsack is the time budget.}

% \todo{Is this needed only for the simulation?}

\section{Flow-Based Solution to the Task Graph}
\label{sec:flow-solution}
In this section, we represent the MRTA problem described in Section~\ref{subsec:ps} over the directed acyclic task graph, as a min-cost network flow problem. By reinterpreting the robot coalitions assigned to tasks as flows between edges in the task graph, we compute a set of reward-maximizing flows $\{f_{ij}\}$ along the edges $(i,j) \in \mc{E}$.
%The directed acyclic graph structure of the robot task graph leads naturally to a formulation of the task allocation problem as a min-cost network flow problem~\cite{Orlin93}, which seeks a set of reward-maximizing flows $\{f_{ij}\}$ along the edges $(i,j) \in \mc{E}$ from the source to the sink node that incur the minimum cost.Cost may be a function of $f_{ij}$, and each edge has a capacity constraint.
%To model robot task allocation, we relax several assumptions from the min-cost flow problem, and thus cannot use traditional approaches: our reward model is non-linear, we allow flow to terminate wherever it is advantageous to do so, and our edges are uncapacitated.

\begin{definition}[Robot Flows]
Let $f_{ij} \in[0, 1]$ represent the population fraction of robots flowing along edge $(i, j)$ \revision{ [Equation \eqref{eq:flow-def}]. The robots represented by flow $f_{ij}$ are assigned to complete task $\mc{T}_i$, followed by task $\mc{T}_j$. According to this definition, the total available flow at the source node $0$ is equal to 1 \eqref{eq:global-flow-constraint}}, where $\mc{N}_i^{out} = \{\mc{T}_j\mid (i, j) \in \mc{E}\}$.

    \noindent\begin{minipage}{0.2\textwidth}
\begin{equation} \label{eq:flow-def}
    \sum_{i \in \Njin} f_{ij} = \frac{C_j}{N}
\end{equation} 
    \end{minipage}%
    \hspace{1.1cm}
    \begin{minipage}{0.2\textwidth}
\begin{equation} \label{eq:global-flow-constraint}
    \sum_{k\in \mc{N}_0^{\mathrm{out}}} f_{0k} \leq 1 
\end{equation}
    \end{minipage}\vskip1em

\end{definition}

 The number of robots flowing out of a node cannot exceed the number of robots that flow into that node. This physical limitation is represented by the following conservation of flow constraint:

\begin{equation} \label{eq:nodewise-flow-constraint}
    \sum_{k \in \Njout} f_{jk} \leq \sum_{i \in \Njin} f_{ij} \quad \forall j \in \mc{M}
\end{equation}
The representation of task assignments with population fractions results in solution methods that are \textit{agnostic to the number of agents in the mission}: larger total coalition sizes do not increase computational complexity. 

\revision{No explicit precedence constraints are necessary in this formulation because precedence relationships are encoded in the graph topology and reward functions.} Note that the inequality in \eqref{eq:nodewise-flow-constraint} allows robots to leave the coalition if necessary. Furthermore, we do not assume any capacity limits on the flow along the task graph edges.
%{\color{red} Dropping two levels of section! Add a sentence on what the following two points are about?}

\subsubsection{Graph Pruning to Satisfy the Makespan Constraint}
As described in Section~\ref{subsec:ps}, the makespan of the mission is constrained, and this must be reflected in the network flow formulation. Towards this end, we prune nodes from the task graph if robots cannot complete them within the makespan budget.
We generate a \textit{makespan graph} $\mc{G}_M = (\Mbar, \mc{E})$ with identical topology to the task graph $\mc{G}_T$.
Each path in the graph has a duration equal to the sum of all node durations in the path.
\revision{Each node $j$ is labeled with a \textit{maximum} (worst-case) finish time $F_j$ corresponding to the longest duration path that terminates at that node.}
This conservative definition is necessary because we assume that robot coalitions must be fully formed to begin a task.
We prune from the graph any node $j$ for which $F_j > T_f$. Therefore, any flow solution over the pruned graph will respect the makespan constraint. 

\subsubsection{Rounding of Population Fraction} Additionally, the flow-based solution deals in the continuous measure of population fraction, whereas real-world task allocation requires discrete quantities of robots in order to compute a task schedule for each robot. In order to translate flow solutions into realistic allocations of robots to tasks, we apply a rounding algorithm. 
% This approximation works well for large quantities of robots, but becomes less accurate as robot population nears 1.
% Accordingly, we translate continuous population fractions into discrete robot quantities via a rounding algorithm.
Iterating over the nodes in the task graph in a topologically sorted manner, we compute a minimum error rounding of the flows along the outgoing edges that respects the flow conservation constraints~\eqref{eq:nodewise-flow-constraint}. This results in a vector of integer flows that can be converted to individual robot schedules.

\subsection{Non-Linear Programming (NLP) Flow Solution}
\label{sec:nlp-solution}
Our primary approach to solving the task allocation problem via network flow is through nonlinear programming, using an off-the-shelf solver \cite{scipy}. The objective is to maximize the task reward sum \eqref{eq:cost} \revision{on the pruned graph} via a set of flows $\{f_{ij} | (i,j) \in \mc{E}\}$ respecting the constraints~\eqref{eq:global-flow-constraint} and~\eqref{eq:nodewise-flow-constraint}. 

\subsection{Greedy One-Step Lookahead Flow Solution}
\label{sec:greedy}
Because we have no solver that guarantees a globally optimal solution to the task graph problem, we develop a greedy one-step lookahead algorithm as a basis of comparison for the NLP solution. The greedy algorithm starts from node 0 and proceeds sequentially through the graph nodes in topological order. For each node $\mc{T}_j$, the greedy algorithm allocates all available flow among the outgoing edges of the node according to which allocation maximizes the rewards of the outgoing neighborhood $\sum_{k \in \mc{N}_j^{out}}r_k$. This local maximum is computed by first taking the best of 50 random samples of the outgoing flow space. We then perform gradient ascent on the reward sum $\sum_{k \in \mc{N}_j^{out}}r_k$ to come to the final values for the outgoing edges of $\mc{T}_j$. This one-step lookahead is myopic but computationally simple. \revision{The gradient ascent step requires task coalition functions and aggregation functions to be differentiable.}

\section{Mixed Integer Non-Linear Programming (MINLP) Solution to the Task Graph}
\label{sec:minlp}
This section describes a mixed integer non-linear programming (MINLP) approach, which leverages the mission model from Section~\ref{sec:mission_model} but differs from the flow-based approach in two crucial ways:
(1)~plans are constructed {\em robot-wise}, generating an ordered schedule of tasks for each robot;
(2)~the robots can switch branches of the task graph.
The objective is to maximize the total reward, as described in Equations~\eqref{eq:reward_dynamics} and~\eqref{eq:cost}, with the coalition size of robots assigned to a task $\mc T_k,\ k\in \mc M$ given by $C_k = \sum_{r\in \mc R}x^r_k$.

\revision{We let non-negative continuous variables $S_k$ and $F_k$ denote the start and the finish times for the task~$\mc{T}_k$.
Additionally, let the binary variable $x_k$ be non-zero when the task $\mc T_k$ is executed by at least one robot, as ensured by constraints~\eqref{eqn:oneTask} below.
The precedence relationships are dictated by the task graph $\mc G_T=(\overline{\mc M}, \mc E)$.
The following constraints \eqref{eqn:precedence} and \eqref{eqn:qij}, with the help of auxiliary binary variables $w_{ij}$, ensure that for an edge $(i, j) \in \mc E$, the preceding task $\mc T_i$ is executed before the dependent task $\mc T_j$, if and only if both $\mc T_i$ and $\mc T_j$ are executed.
	Furthermore, for each executed task, we need to ensure that the duration constraints \eqref{eqn:duration} are enforced.}
\begin{align}
	&x_k \geq x_k^a, \, \forall a\in \mc R \text{ and } x_k \leq \sum_{r\in \mc R}x_k^r, \quad \forall k \in \mc M\label{eqn:oneTask}\\
	&w_{ij} (S_j - F_i) \geq 0,\; w_{ij}\geq x_i + x_j -1,\label{eqn:precedence}\\
	&w_{ij}\leq x_i,\;w_{ij} \leq x_j,\text{ and } w_{ij}\in\{0,1\},\quad \forall (i, j) \in \mc E\label{eqn:qij}\\
	&x_k \left(F_k - S_k\right) \geq x_k d_k, \quad \forall k \in \mc M\label{eqn:duration}
\end{align}

The rest of the constraints follow from the formulation by Nunes et al.~\cite{NUNES2017}.
Binary variables $z_k^r$ denote if $k$ is the last task executed by the robot~$r$, and binary variables $o_{kk'}^r$ denote if robot~$r$ executes a task $k'\in \mc M$ immediately after task $k$.
\begin{align*}
	\sum_{k\in \overline{\mc M}} o^r_{kk'} &= x^r_{k'},  &&\forall r\in \mc R,\ \forall k' \in \mc M \\
	\sum_{k'\in \mc M} o^r_{kk'} + z^r_k &= x^r_{k},  &&\forall r\in \mc R,\ \forall k \in \overline{\mc M} \\
	\sum_{k\in \overline{\mc M}} x^r_k\, z^r_k &= 1,\;x^r_0=1  &&\forall r\in \mc R \\
	o^r_{kk'}\left(S_{k'} - F_k\right) &\geq 0,  &&\forall k \in \overline{\mc M},\ \forall k' \in \mc M \\
	F_k &\leq T_f,  &&\forall k\in \mc M
\end{align*}

Given the computational complexity of solving a mixed-integer problem with non-linear constraints, this approach is more applicable to instances with sparser precedence relationships and fewer robots.
\revision{Nevertheless, as solving the MILNP provides an optimal solution when solved to convergence, it serves as a benchmark for heuristic solvers.}
The MINLP formulation admits a trivial solution corresponding to none of the tasks being executed.
Thus, when solved using methods such as branch-and-bound, it always maintains a set of feasible solutions, and the solver can be terminated {\em anytime} to obtain a non-optimal feasible solution.
% However, the formulation comprises quadratic constraints and can take a long time to compute an optimal solution.

% \begin{figure}
% 	\centering 
% 	\includegraphics[trim={0.0cm 0.0cm 0.0cm 0.0cm},clip,width=\linewidth]{figures/graph_generator_placeholder.png}
% 	\caption{\label{fig:graph_generator} [PLACEHOLDER: figure demonstrating the breadth of random graph generation?]}
% \end{figure}
 
\section{Experimental Evaluation}
\label{sec:experiments}
We evaluate the three algorithms on two simulated sets of scenarios. The first generates random task graphs given a target number of tasks. The second is a block dry-stacking scenario that represents the mission structures present in autonomous construction settings. \revision{All experiments were run on a desktop computer with an 8-core Intel Core i7-9700 CPU.
Our implementation uses the PySCIPOpt optimization suite~\cite{MaherMiltenbergerPedrosoetal.2016, BestuzhevaEtal2021OO} as the solver for the MINLP approach and SciPy optimize~\cite{scipy} for the flow-based NLP approach.}

\subsection{Random Graph Generator Platform and Experiments}
We developed a random graph generator that generates graphs with random topologies and reward functions, allowing us to explore the mission specification space broadly. We specify a target number of tasks, and randomly generate coalition and influence functions. We sample from polynomial, sub-linear, and sigmoid functions (a smoothed approximation of a step function) with randomized parameters.
We sample randomly between using a sum and product as $\ddag$, which combines the coalition and influence function outputs. For these experiments, we use the sum operation as our influence aggregation function $\alpha_j$ for each task $\mc{T}_j$.

\subsubsection{Experiment 1: Number of Tasks}

In this experiment, we evaluate the impact of the number of tasks in the mission on the relative performance of the three solution methods. For each number of tasks, we conduct ten trials consisting of unique randomly generated task graphs. For this experiment, we use 4 agents, a makespan constraint $T_f = 0.6\sum_{j \in \mathcal{M}}d_j$, and limit computation time to 10 minutes per trial.

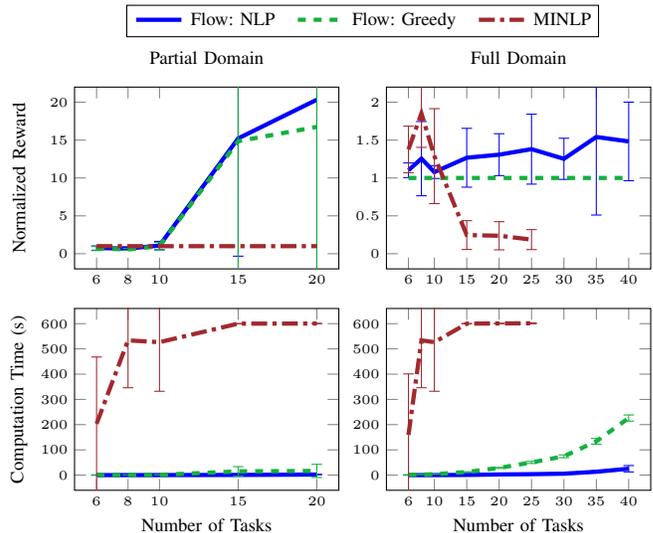
\begin{figure}
    \centering
    \definecolor{mLightGreen}{HTML}{14B03D}
\definecolor{mDarkRed}{HTML}{a2282f}
\tikzstyle{mydashdot}=[dash pattern=on 6pt off 2pt on \the\pgflinewidth off 2pt]
\usetikzlibrary{matrix}
\usepgfplotslibrary{groupplots}
\pgfplotsset{every tick label/.append style={font=\tiny}}
\pgfplotsset{ylabsh/.style={every axis y label/.style={at={(0,0.5)}, xshift=#1, rotate=90}}}  

\begin{tikzpicture}
	\scriptsize
	\begin{groupplot}[group style={
			group name=taskplot,
		vertical sep=15pt,
		horizontal sep=18pt,
		group size= 2 by 4},
		height=4cm,
		width=0.59\linewidth,
		ylabel near ticks,
		ylabsh=-3.0em,
		every axis plot/.append style={ultra thick},
		enlarge y limits,
		enlarge x limits,
		]
		\nextgroupplot[
		title=Partial Domain,
		ylabel={Normalized Reward},
		ymin=0,
		ymax=20,
		xtick={6,8,10,15,20},
		ytick={0,5,10,15,20},
		]
		\addplot[
		blue,
		error bars/.cd,
		y dir=both,
		y explicit,
		]
		coordinates {
			(6,0.7476727145) +- (0,0.2850967917)
			(8,0.6851674580) +- (0,0.208032099)
			(10,1.079841979) +- (0,0.5548580907)
			(15,15.22405253) +- (0,15.54604511)
			(20,20.33659090) +- (0,30.1613607)
		};
		\label{plts:tasks:flow_nlp}
		\addplot[
		mLightGreen,
		dashed,
		error bars/.cd,
		y dir=both,
		y explicit,
		error bar style={solid},
		]
		coordinates {
			(6,0.6820733006) +- (0,0.2664457739)
			(8,0.5716402778) +- (0,0.1374782218)
			(10,0.9980552294) +- (0,0.5084652094)
			(15,14.85934384) +- (0,18.74224624)
			(20,16.74179163) +- (0,25.98545479)
		};
		\label{plts:tasks:flow_greedy}
		\addplot[
		mDarkRed,
		mydashdot,
		error bars/.cd,
		y dir=both,
		y explicit,
		error bar style={solid},
		]
		coordinates {
			(6,1)
			(8,1)
			(10,1)
			(15,1)
			(20,1)
		};
		\label{plts:tasks:minlp}
		\nextgroupplot[
		title=Full Domain,
		ymin=0,
		ymax=2,
		xtick={6,10,15,20,25,30,35,40},
		ytick={0,0.5,1.0,1.5,2.0},
		]
		\addplot[
		blue,
		error bars/.cd,
		y dir=both,
		y explicit,
		]
		coordinates {
			(6,1.102012612) +- (0,0.09710291465)
			(8,1.255297827) +- (0,0.4901981637)
			(10,1.0778136) +- (0,0.08456967601)
			(15,1.267152193) +- (0,0.3883309599)
			(20,1.307925123) +- (0,0.2767126582)
			(25,1.380677498) +- (0,0.4617766767)
			(30,1.253010562) +- (0,0.2717426982)
			(35,1.541391982) +- (0,1.03064376)
			(40,1.482796638) +- (0,0.5188277107)
		};
		\addplot[
		mLightGreen,
		dashed,
		error bars/.cd,
		y dir=both,
		y explicit,
		error bar style={solid},
		]
		coordinates {
			(6,1)
			(8,1)
			(10,1)
			(15,1)
			(20,1)
			(25,1)
			(30,1)
			(35,1)
			(40,1)
		};
		\addplot[
		mDarkRed,
		mydashdot,
		error bars/.cd,
		y dir=both,
		y explicit,
		error bar style={solid},
		]
		coordinates {
			(6,1.376914101) +- (0,0.3070793986)
			(8,1.853693332) +- (0,0.449513763)
			(10,1.288519949) +- (0,0.6268963864)
			(15,0.2464919801) +- (0,0.1900516635)
			(20,0.2374847911) +- (0,0.1858953404)
			(25,0.1866407244) +- (0,0.1319194971)
		};
		\nextgroupplot[
		ymin=0,
		ymax=600,
		xtick={6,8,10,15,20},
		ytick={0,100,200,300,400,500,600},
		xlabel={Number of Tasks},
		ylabel={Computation Time (s)}
		]
		\addplot[
		blue,
		error bars/.cd,
		y dir=both,
		y explicit,
		]
		coordinates {
			(6,0.04270567894) +- (0,0.02297552665)
			(8,0.07417503993) +- (0,0.03083867518)
			(10,0.150190264) +- (0,0.05762542032)
			(15,0.6987413168) +- (0,0.4867874711)
			(20,2.260855397) +- (0,0.6238764405)
		};
		\addplot[
		mLightGreen,
		dashed,
		error bars/.cd,
		y dir=both,
		y explicit,
		error bar style={solid},
		]
		coordinates {
			(6,0.6820733006) +- (0,0.2664457739)
			(8,0.5716402778) +- (0,0.1374782218)
			(10,0.9980552294) +- (0,0.5084652094)
			(15,14.85934384) +- (0,18.74224624)
			(20,16.74179163) +- (0,25.98545479)
		};
		\addplot[
		mDarkRed,
		mydashdot,
		error bars/.cd,
		y dir=both,
		y explicit,
		error bar style={solid},
		]
		coordinates {
			(6,203.2845593) +- (0,264.5867665)
			(8,533.6892898) +- (0,187.5831867)
			(10,526.5333105) +- (0,194.3983954)
			(15,600.0102864) +- (0,0.01931346256)
			(20,600.4096871) +- (0,0.5369724171)
		};

		\nextgroupplot[
		ymin=0,
		ymax=600,
		xtick={6,10,15,20,25,30,35,40},
		ytick={0,100,200,300,400,500,600},
		xlabel={Number of Tasks},
		]
		\addplot[
		blue,
		error bars/.cd,
		y dir=both,
		y explicit,
		]
		coordinates {
			(6,0.04227670034) +- (0,0.02418030927)
			(8,0.07417503993) +- (0,0.03083867518)
			(10,0.150190264) +- (0,0.05762542032)
			(15,0.6987413168) +- (0,0.4867874711)
			(20,2.636538911) +- (0,0.9246638163)
			(25,3.766391188) +- (0,1.314147213)
			(30,5.712127995) +- (0,1.94728771)
			(35,13.32392333) +- (0,3.691170741)
			(40,25.31834844) +- (0,12.65423681)
		};
		\addplot[
		mLightGreen,
		dashed,
		error bars/.cd,
		y dir=both,
		y explicit,
		error bar style={solid},
		]
		coordinates {
			(6,1.027963188) +- (0,0.3086112777)
			(8,2.100535472) +- (0,0.3841814074)
			(10,3.634620398) +- (0,0.50581861)
			(15,12.23006787) +- (0,1.130411674)
			(20,28.59756718) +- (0,2.077900782)
			(25,50.36679554) +- (0,4.130514145)
			(30,74.52776892) +- (0,5.639944524)
			(35,133.9933318) +- (0,11.2876844)
			(40,226.0408998) +- (0,12.20468468)
		};
		\addplot[
		mDarkRed,
		mydashdot,
		error bars/.cd,
		y dir=both,
		y explicit,
		error bar style={solid},
		]
		coordinates {
			(6,159.2049692) +- (0,241.5667947)
			(8,533.6892898) +- (0,187.5831867)
			(10,526.5333105) +- (0,194.3983954)
			(15,600.0102864) +- (0,0.01931346256)
			(20,600.765846) +- (0,0.9469208314)
			(25,601.5004564) +- (0,1.980313241)
		};
	\end{groupplot}
	% \path (taskplot c1r1.outer north west)% plot in column 1 row 1
	%       -- node[anchor=south,rotate=90] {throughput}% label midway
	%       (taskplot c1r4.outer south west)% plot in column 1 row 4
	% ;
	% legend
	\path (taskplot c1r1.north west|-current bounding box.north)--
	coordinate(legendpos)
	(taskplot c2r1.north east|-current bounding box.north);
	\matrix[
	matrix of nodes,
	anchor=south,
	draw,
	inner sep=0.2em,
	draw
	]at([yshift=1ex]legendpos)
	{
		\ref{plts:tasks:flow_nlp}& Flow: NLP&[5pt]
		\ref{plts:tasks:flow_greedy}& Flow: Greedy&[5pt]
	\ref{plts:tasks:minlp}& MINLP\\};
\end{tikzpicture}
	\caption{\label{fig:n_tasks} \revision{Experiment 1 tests planner performance as mission size (number of tasks) is increased---we measure solution reward (top) and computation time (bottom). The left plots depict the portion of the domain where MINLP computes a non-trivial solution, normalized by MINLP reward. The right plots depict the full domain, normalized by the greedy approach's reward. As mission size grows, the flow-based approaches outperform the MINLP approach until the MINLP fails to produce non-trivial solutions.} \vspace{-0.5cm}}
\end{figure}

In Figure \ref{fig:n_tasks} (left), we examine the portion of the domain in which the MINLP solution consistently computes a non-trivial feasible solution: $\leq 20$ tasks. We graph the mean of rewards normalized by the MINLP reward for each trial, e.g., if the reward from the flow NLP solution is 20\% higher than the reward from the MINLP solution, we report $1.2$.

In Figure \ref{fig:n_tasks} (right), we show the full domain we tested, normalize the results by the \textit{greedy} flow solution, and graph the MINLP results only when feasible and non-trivial. On average over all trials in the full domain, the flow-based NLP approach outperforms the greedy approach by 30\% and takes on average 10\% the time of the greedy approach to compute a solution. 
%\begin{figure}
%	\centering 
%	%\includegraphics[trim={19.8cm 1.1cm 4.6cm 5.5cm},clip,width=\linewidth]{figures/fig1_screenshot.png}
%    \includegraphics[trim={0cm 0cm 0cm 0cm},clip,width=\linewidth]{figures/fig1_9_14_labeled.png}
%	\caption{\label{fig:n_tasks} Experiment 1 demonstrating the impact of the number of tasks in the mission on the performance of the three methods---we measure solution reward (top) and computation time (bottom). The left plots depict the portion of the domain where MINLP converges to a non-trivial solution, normalized by MINLP reward. The right plots depict the full domain, normalized by the reward of the solution obtained using the greedy flow approach.}
%\end{figure}

\subsubsection{Experiment 2: Number of Agents}
\begin{figure}
	\centering 
	\begin{tikzpicture}
	\scriptsize
	\begin{groupplot}[group style={
			group name=robotplot,
		vertical sep=15pt,
		horizontal sep=27pt,
		group size= 2 by 4},
		height=4cm,
		width=0.59\linewidth,
		ylabel near ticks,
		every axis plot/.append style={ultra thick},
		enlarge y limits,
		enlarge x limits,
		]
		\nextgroupplot[
		ylabel={Normalized Reward},
		xlabel={Number of Agents},
		ymin=0,
		ymax=20,
		xtick={2,4,6,10,15,20},
		ytick={0,5,10,15,20},
		ylabsh=-2.3em,
		]
		\addplot[
		blue,
		error bars/.cd,
		y dir=both,
		y explicit,
		]
		coordinates {
			(2,0.775185901) +- (0,0.3009462926)
			(4,5.219018955) +- (0,7.522221525)
			(6,3.658508517) +- (0,6.808649718)
			(10,2.997137111) +- (0,3.47232613)
			(15,17.30902005) +- (0,21.1223129)
			(20,11.03597902) +- (0,10.63936932)
		};
		\label{plts:robot:flow_nlp}
		\addplot[
		mLightGreen,
		dashed,
		error bars/.cd,
		y dir=both,
		y explicit,
		error bar style={solid},
		]
		coordinates {
			(2,0.7390198678) +- (0,0.2890756783)
			(4,4.905897163) +- (0,7.415390773)
			(6,3.255932939) +- (0,6.924113608)
			(10,3.348287276) +- (0,3.222582309)
			(15,16.56642546) +- (0,20.29789247)
			(20,8.890838786) +- (0,8.374751345)
		};
		\label{plts:robot:flow_greedy}
		\addplot[
		mDarkRed,
		mydashdot,
		error bars/.cd,
		y dir=both,
		y explicit,
		error bar style={solid},
		]
		coordinates {
			(2,1)
			(4,1)
			(6,1)
			(10,1)
			(15,1)
			(20,1)
		};
		\label{plts:robot:minlp}
		\nextgroupplot[
		ylabel={Computation Time (s)},
		xlabel={Number of Agents},
		ymin=0,
		ymax=600,
		xtick={2,4,6,10,15,20},
		ytick={0,100,200,300,400,500,600},
		ylabsh=-2.6em,
		]
		\addplot[
		blue,
		error bars/.cd,
		y dir=both,
		y explicit,
		]
		coordinates {
			(2,0.1815769196) +- (0,0.07265819741)
			(4,0.1515746911) +- (0,0.05692403011)
			(6,0.1754524708) +- (0,0.08819290968)
			(10,0.1360272037) +- (0,0.03683310284)
			(15,0.1221920649) +- (0,0.003767847622)
			(20,0.167237252) +- (0,0.07085353013)
		};
		\addplot[
		mLightGreen,
		dashed,
		error bars/.cd,
		y dir=both,
		y explicit,
		error bar style={solid},
		]
		coordinates {
			(2,4.011505532) +- (0,0.2604542445)
			(4,3.778797838) +- (0,0.6321676832)
			(6,3.574668074) +- (0,0.526622167)
			(10,3.573601007) +- (0,0.4039821598)
			(15,3.440777779) +- (0,0.07540720283)
			(20,3.593786895) +- (0,0.3076087716)
		};
		\addplot[
		mDarkRed,
		mydashdot,
		error bars/.cd,
		y dir=both,
		y explicit,
		error bar style={solid},
		]
		coordinates {
			(2,476.2564244) +- (0,202.7976925)
			(4,600.013053) +- (0,0.03094763533)
			(6,600.0052922) +- (0,0.005019675416)
			(10,600.0847335) +- (0,0.2312592591)
			(15,600.1522795) +- (0,0.1028297097)
			(20,601.0547325) +- (0,1.245827067)
		};
	\end{groupplot}
	% \path (robotplot c1r1.outer north west)% plot in column 1 row 1
	%       -- node[anchor=south,rotate=90] {throughput}% label midway
	%       (robotplot c1r4.outer south west)% plot in column 1 row 4
	% ;
	% legend
	\path (robotplot c1r1.north west|-current bounding box.north)--
	coordinate(legendpos)
	(robotplot c2r1.north east|-current bounding box.north);
	\matrix[
	matrix of nodes,
	anchor=south,
	draw,
	inner sep=0.2em,
	draw
	]at([yshift=1ex]legendpos)
	{
		\ref{plts:robot:flow_nlp}& Flow: NLP&[5pt]
		\ref{plts:robot:flow_greedy}& Flow: Greedy&[5pt]
	\ref{plts:robot:minlp}& MINLP\\};
\end{tikzpicture}
    \vspace{-0.5cm}
	\caption{\label{fig:n_agents_exp} Experiment 2 demonstrates the impact of the number of agents in the coalition on the performance of the three solution methods. We plot reward (left) and computation time (right).
	The rewards are normalized by the MINLP solution.
	As the number of agents increases, the MINLP problem size grows, whereas the flow-based solutions only get more accurate, due to decreased rounding error.\vspace{-0.1cm}}
\end{figure}
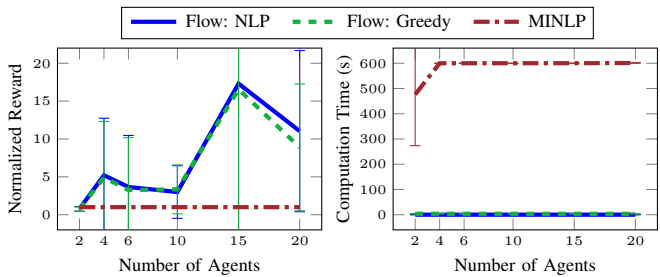

Experiment 2 evaluates the relative performance of the solution methods as the number of agents varies. We test with total agent populations of \{2,4,6,10,15,20\}, and conduct 10 trials with unique randomly generated task graphs for each value. For this experiment, we use 10 tasks, a makespan constraint $T_f = 0.6\sum_{j \in \mathcal{M}}d_j$, and limit computation time to 10 minutes per trial. In Figure \ref{fig:n_agents_exp}, we plot the rewards normalized by the MINLP reward, and the computation time for each method.

\subsubsection{Experiment 3: Makespan and Number of Tasks}
\begin{figure}
	\centering 
	\includegraphics[trim={0.2cm 0.0cm 0.2cm 0.0cm},clip,width=\linewidth]{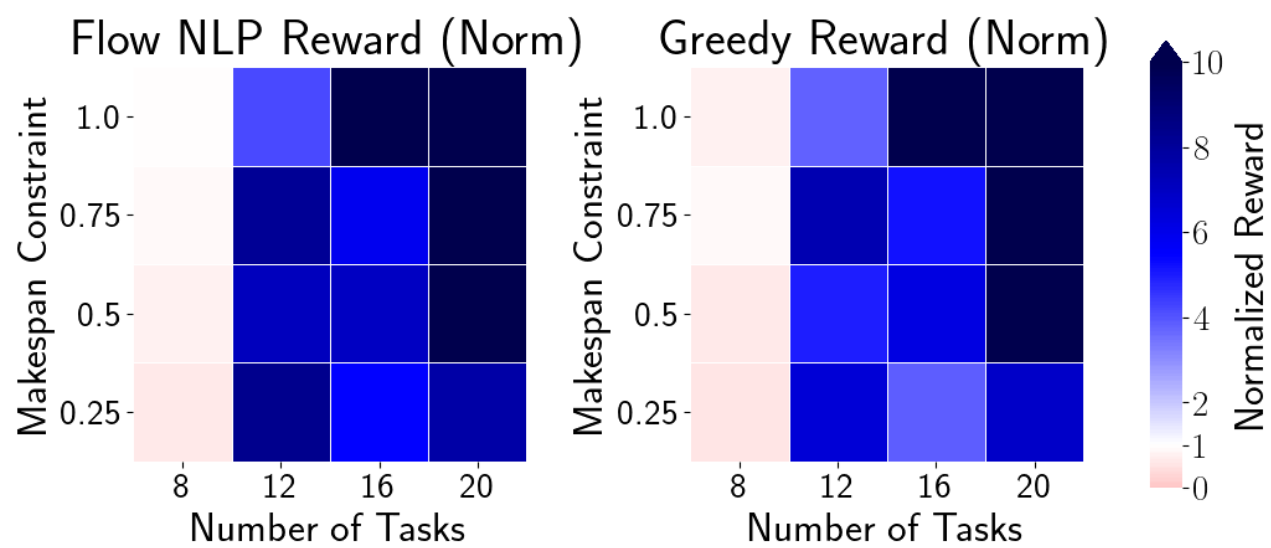}
	\caption{\label{fig:makespan_exp} \edit{The reward accrued by the NLP flow approach (left) and the greedy flow approach (right) for varying makespan constraint and number of tasks.
 The rewards are normalized by the MINLP reward. The flow-based approaches perform worse than the MINLP on small problems with a tight makespan constraint, but outperform on problems larger than 8 tasks.  The MINLP approach fails to compute a non-trivial solution on larger problems than those shown here.}\vspace{-0.0cm}}
\end{figure}

In Experiment 3, we examine the impact of both the makespan constraint and the number of tasks on the relative performance of the solution methods. In this experiment, we examine \textit{only the portion of the domain in which the MINLP approach computes non-trivial solutions.} We formulate the makespan constraint for randomly generated graphs as a proportion of the total duration of all tasks, $\mu \sum_{j \in \mc{M}}d_j$. We evaluate $\mu=\{0.25,0.5,0.75,1.0\}$ and $M=\{8,12,16,20\}$ tasks and conduct 10 trials at each pair of values. For this experiment, we use 4 agents and limit computation time to 10 minutes per trial. We show the NLP and greedy flow solution results normalized by the MINLP results in Figure~\ref{fig:makespan_exp}.

%\caption{\label{fig:makespan_exp} \edit{The reward accrued by the NLP flow approach (left) and the greedy flow approach (right) and normalized by the MINLP reward, shown as the makespan constraint and number of tasks in the problem are varied. The flow-based approaches perform worse than MINLP on small problems with a tight makespan constraint, but outperform on problems larger than 8 tasks.  MINLP fails to compute a non-trivial solution on larger problems than those shown here.}\vspace{-0.0cm}}

\subsection{Autonomous Construction Platform and Experiments}
\revision{In order to evaluate our system on task graphs derived from realistic mission structures, we created a dry-stacking problem generator that is applicable to autonomous construction scenarios.} Given a tower base width and a number of layers, we create a layered task graph representing the tower, where each block placement task is a node and block heights and widths are randomized. Blocks in adjacent layers have precedence relationships. Each block task's coalition function is a sigmoid function structured such that heavier blocks require larger coalitions to place. 
%When executing construction plans, larger robot coalitions for blocks result in less uncertainty in the reward as well. Construction makespan and reward can meaningfully be compared across our task graph solution methods.

\begin{figure}
	\centering 
	\includegraphics[trim={0.0cm 0.0cm 0.0cm 0.0cm},clip,width=\linewidth]{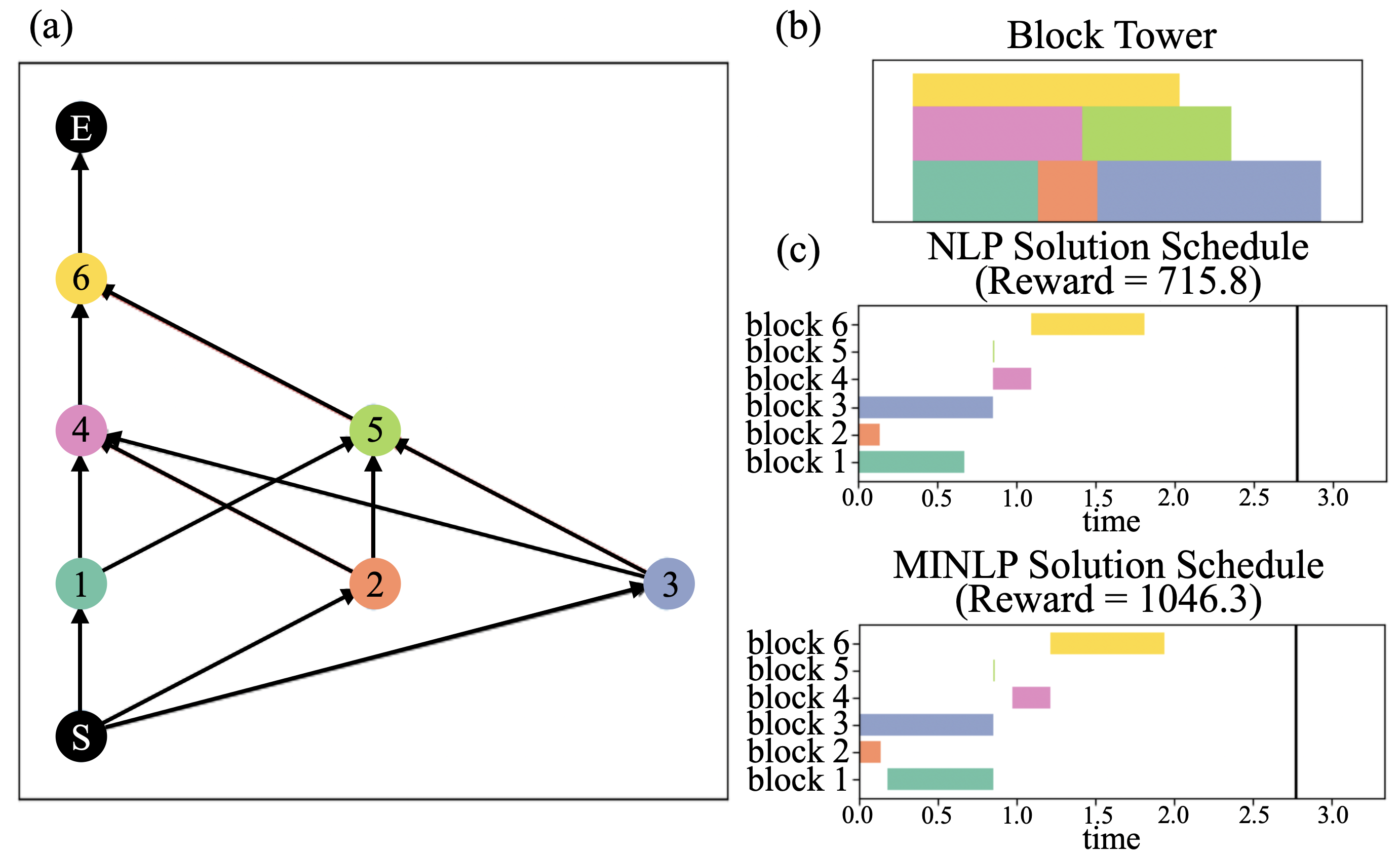}
	\caption{\label{fig:construction_env} (a)~Generated autonomous construction graph. Start and end nodes are labeled ``S'' and ``E''. (b) Corresponding block tower. (c) Construction schedules for NLP (flow) and MINLP solutions. The makespan constraint is set to 2.79, shown as the vertical black line. \revision{In this example, the MINLP approach is able to allocate a group of agents to perform task 2 followed by task 1, while the flow-based approaches must divide agents among the edges (0,1), (0,2), and (0,3). }\vspace{-0.1cm}}
\end{figure}

In Table \ref{tab:autonomous-construction}, we show the results of an experiment conducted with the autonomous construction testbed. We generate block towers with $\{8,12,16,20\}$ blocks, with 5 random trials for each value. We test all three solution methods, and report the rewards of the NLP flow and greedy flow solutions, normalized by the MINLP reward on each trial.
Figure \ref{fig:construction_env} illustrates the block tower, corresponding task graph, and the solution schedules of the NLP flow and MINLP solutions for one autonomous construction example. 

\revision{As this experiment demonstrates, the MINLP approach solves small problems optimally but performs poorly on large problems. The flow-based NLP approach provides high-quality solutions albeit non-optimal but boasts orders of magnitude faster computation. This enables solving much larger problems and makes it well-suited to online applications, which we will explore in future work.}
\begin{table}
    \begin{center}
    \caption{Autonomous Construction Experiment Results\label{tab:autonomous-construction}\vspace{0.3cm}}
        \begin{tabular}{|c|c|c|}
            \hline
            \textbf{No. Tasks} & \textbf{Norm. NLP} & \textbf{Norm. Greedy} \\
            \hline
            8 & 0.944 & 0.656\\
            12 & 2.299 & 1.644\\
            16 & 1.874 & 1.311\\
            20 & 1.843 & 1.117\\
            \hline
        \end{tabular}
        \vspace{-0.0cm}
    \end{center}
\end{table}

\section{Conclusion}
In this paper, we presented a framework that models coordination and cooperation in missions consisting of inter-dependent multi-robot tasks. Our task graph formulation, coupled with expressive task precedence relationships and reward-based coalition models, enabled the automatic generation of task plans in such complex missions. We presented flow-based solution approaches that
use a coalition fraction to scale to problems with any number of agents.
These approaches perform faster and better than the mixed integer approach on large problem sizes by leveraging the fundamental graph structure of the task allocation problem.
\balance
\bibliographystyle{IEEEtran}
\bibliography{refs}

\begin{thebibliography}{10}
\providecommand{\url}[1]{#1}
\csname url@samestyle\endcsname
\providecommand{\newblock}{\relax}
\providecommand{\bibinfo}[2]{#2}
\providecommand{\BIBentrySTDinterwordspacing}{\spaceskip=0pt\relax}
\providecommand{\BIBentryALTinterwordstretchfactor}{4}
\providecommand{\BIBentryALTinterwordspacing}{\spaceskip=\fontdimen2\font plus
\BIBentryALTinterwordstretchfactor\fontdimen3\font minus
  \fontdimen4\font\relax}
\providecommand{\BIBforeignlanguage}[2]{{%
\expandafter\ifx\csname l@#1\endcsname\relax
\typeout{** WARNING: IEEEtran.bst: No hyphenation pattern has been}%
\typeout{** loaded for the language `#1'. Using the pattern for}%
\typeout{** the default language instead.}%
\else
\language=\csname l@#1\endcsname
\fi
#2}}
\providecommand{\BIBdecl}{\relax}
\BIBdecl

\bibitem{gerkey2004formal}
B.~P. Gerkey and M.~J. Matari{\'c}, ``A formal analysis and taxonomy of task
  allocation in multi-robot systems,'' \emph{The International Journal of
  Robotics Research}, vol.~23, no.~9, pp. 939--954, 2004.

\bibitem{BurkardDM12}
R.~Burkard, M.~Dell'Amico, and S.~Martello, \emph{Assignment Problems}.\hskip
  1em plus 0.5em minus 0.4em\relax SIAM, 2012.

\bibitem{knepper2013ikeabot}
R.~A. Knepper, T.~Layton, J.~Romanishin, and D.~Rus, ``Ikeabot: An autonomous
  multi-robot coordinated furniture assembly system,'' in \emph{IEEE
  International Conference on Robotics and Automation (ICRA)}.\hskip 1em plus
  0.5em minus 0.4em\relax IEEE, 2013, pp. 855--862.

\bibitem{mao2021research}
W.~Mao, Z.~Liu, H.~Liu, F.~Yang, and M.~Wang, ``Research progress on
  synergistic technologies of agricultural multi-robots,'' \emph{Applied
  Sciences}, vol.~11, no.~4, p. 1448, 2021.

\bibitem{deng2019compiler}
Y.~Deng, Y.~Hua, N.~Napp, and K.~Petersen, ``A compiler for scalable
  construction by the termes robot collective,'' \emph{Robotics and Autonomous
  Systems}, vol. 121, p. 103240, 2019.

\bibitem{dogar2019multi}
M.~Dogar, A.~Spielberg, S.~Baker, and D.~Rus, ``Multi-robot grasp planning for
  sequential assembly operations,'' \emph{Autonomous Robots}, vol.~43, no.~3,
  pp. 649--664, 2019.

\bibitem{gombolay2013fast}
M.~C. Gombolay, R.~J. Wilcox, and J.~A. Shah, ``Fast scheduling of robot teams
  performing tasks with temporospatial constraints,'' \emph{IEEE Transactions
  on Robotics}, vol.~34, no.~1, pp. 220--239, 2018.

\bibitem{NUNES2017}
E.~Nunes, M.~Manner, H.~Mitiche, and M.~Gini, ``A taxonomy for task allocation
  problems with temporal and ordering constraints,'' \emph{Robotics and
  Autonomous Systems}, vol.~90, pp. 55--70, 2017, special Issue on New Research
  Frontiers for Intelligent Autonomous Systems.

\bibitem{kolen1987vehicle}
A.~W. Kolen, A.~Rinnooy~Kan, and H.~W. Trienekens, ``Vehicle routing with time
  windows,'' \emph{Operations Research}, vol.~35, no.~2, pp. 266--273, 1987.

\bibitem{bredstrom2008combined}
D.~Bredstr{\"o}m and M.~R{\"o}nnqvist, ``Combined vehicle routing and
  scheduling with temporal precedence and synchronization constraints,''
  \emph{European Journal of Operational Research}, vol. 191, no.~1, pp. 19--31,
  2008.

\bibitem{brafman2014distributed}
R.~Brafman and U.~Zoran, ``Distributed heuristic forward search for multi-agent
  systems,'' in \emph{ICAPS DMAP workshop}, 2014, pp. 1--6.

\bibitem{shekhar2020signaling}
S.~Shekhar, R.~I. Brafman, and G.~Shani, ``Signaling in contingent multi-agent
  planning,'' in \emph{Workshop on Epistemic Planning (EpiP)}, 2020.

\bibitem{TERESHCHUK2021102154}
V.~Tereshchuk, N.~Bykov, S.~Pedigo, S.~Devasia, and A.~G. Banerjee, ``A
  scheduling method for multi-robot assembly of aircraft structures with soft
  task precedence constraints,'' \emph{Robotics and Computer-Integrated
  Manufacturing}, vol.~71, p. 102154, 2021.

\bibitem{smith2019real}
A.~J. Smith, G.~Best, J.~Yu, and G.~A. Hollinger, ``Real-time distributed
  non-myopic task selection for heterogeneous robotic teams,'' \emph{Autonomous
  Robots}, vol.~43, no.~3, pp. 789--811, 2019.

\bibitem{ponda2010decentralized}
S.~Ponda, J.~Redding, H.-L. Choi, J.~P. How, M.~Vavrina, and J.~Vian,
  ``Decentralized planning for complex missions with dynamic communication
  constraints,'' in \emph{Proceedings of the 2010 American Control
  Conference}.\hskip 1em plus 0.5em minus 0.4em\relax IEEE, 2010, pp.
  3998--4003.

\bibitem{wang2022heterogeneous}
Z.~Wang, C.~Liu, and M.~Gombolay, ``Heterogeneous graph attention networks for
  scalable multi-robot scheduling with temporospatial constraints,''
  \emph{Autonomous Robots}, vol.~46, no.~1, pp. 249--268, 2022.

\bibitem{Korsah2013}
G.~A. Korsah, M.~B. Dias, and A.~Stentz, ``A comprehensive taxonomy for
  multi-robot task allocation background,'' \emph{The International Journal of
  Robotics Research}, pp. 1--29, 2013.

\bibitem{seenu2020review}
N.~Seenu, K.~C. RM, M.~Ramya, and M.~N. Janardhanan, ``Review on
  state-of-the-art dynamic task allocation strategies for multiple-robot
  systems,'' \emph{Industrial Robot: The International Journal of Robotics
  Research and Application}, vol.~47, no.~6, pp. 929--942, 2020.

\bibitem{prorok2017impact}
A.~Prorok, M.~A. Hsieh, and V.~Kumar, ``The impact of diversity on optimal
  control policies for heterogeneous robot swarms,'' \emph{IEEE Transactions on
  Robotics}, vol.~33, no.~2, pp. 346--358, 2017.

\bibitem{mayya2021resilient}
S.~Mayya, D.~S. D’antonio, D.~Saldaña, and V.~Kumar, ``Resilient task
  allocation in heterogeneous multi-robot systems,'' \emph{IEEE Robotics and
  Automation Letters}, vol.~6, no.~2, pp. 1327--1334, 2021.

\bibitem{messing2022grstaps}
A.~Messing, G.~Neville, S.~Chernova, S.~Hutchinson, and H.~Ravichandar,
  ``{GRSTAPS}: Graphically recursive simultaneous task allocation, planning,
  and scheduling,'' \emph{The International Journal of Robotics Research},
  vol.~41, no.~2, pp. 232--256, 2022.

\bibitem{capezzuto2020anytime}
L.~Capezzuto, D.~Tarapore, and S.~Ramchurn, ``Anytime and efficient coalition
  formation with spatial and temporal constraints,'' in \emph{Multi-Agent
  Systems and Agreement Technologies}, N.~Bassiliades, G.~Chalkiadakis, and
  D.~de~Jonge, Eds.\hskip 1em plus 0.5em minus 0.4em\relax Cham: Springer
  International Publishing, 2020, pp. 589--606.

\bibitem{ramchurn2010coalition}
S.~D. Ramchurn, M.~Polukarov, A.~Farinelli, C.~Truong, and N.~R. Jennings,
  ``Coalition formation with spatial and temporal constraints,'' in
  \emph{Proceedings of the 9th International Conference on Autonomous Agents
  and Multiagent Systems: Volume 3}, ser. AAMAS '10.\hskip 1em plus 0.5em minus
  0.4em\relax Richland, SC: International Foundation for Autonomous Agents and
  Multiagent Systems, 2010, pp. 1181--1188.

\bibitem{korsah2012xbots}
G.~A. Korsah, B.~Kannan, B.~Browning, A.~Stentz, and M.~B. Dias, ``x{B}ots: An
  approach to generating and executing optimal multi-robot plans with
  cross-schedule dependencies,'' in \emph{IEEE International Conference on
  Robotics and Automation}, 2012, pp. 115--122.

\bibitem{prorok2021beyond}
A.~Prorok, M.~Malencia, L.~Carlone, G.~S. Sukhatme, B.~M. Sadler, and V.~Kumar,
  ``Beyond robustness: A taxonomy of approaches towards resilient multi-robot
  systems,'' \emph{arXiv preprint arXiv:2109.12343}, 2021.

\bibitem{dutta2019one}
A.~Dutta and A.~Asaithambi, ``One-to-many bipartite matching based coalition
  formation for multi-robot task allocation,'' in \emph{International
  Conference on Robotics and Automation (ICRA)}.\hskip 1em plus 0.5em minus
  0.4em\relax IEEE, 2019, pp. 2181--2187.

\bibitem{zitouni2019fa}
F.~Zitouni, R.~Maamri, and S.~Harous, ``{FA--QABC--MRTA}: a solution for
  solving the multi-robot task allocation problem,'' \emph{Intelligent Service
  Robotics}, vol.~12, no.~4, pp. 407--418, 2019.

\bibitem{gens1980complexity}
G.~Gens and E.~Levner, ``Complexity of approximation algorithms for
  combinatorial problems: a survey,'' \emph{ACM SIGACT News}, vol.~12, no.~3,
  pp. 52--65, 1980.

\bibitem{scipy}
\BIBentryALTinterwordspacing
Scipy, ``Docs.scipy.org.'' 2022, date accessed: 2022-05-10. [Online].
  Available:
  \url{https://docs.scipy.org/doc/scipy/reference/generated/scipy.optimize.minimize.html\#scipy-optimize-minimize}
\BIBentrySTDinterwordspacing

\bibitem{MaherMiltenbergerPedrosoetal.2016}
S.~J. Maher, M.~Miltenberger, J.~P. Pedroso, D.~Rehfeldt, R.~Schwarz, and
  F.~Serrano, ``Pyscipopt: Mathematical programming in python with the scip
  optimization suite,'' in \emph{Mathematical Software - ICMS 2016}, vol. 9725,
  2016, pp. 301 -- 307.

\bibitem{BestuzhevaEtal2021OO}
\BIBentryALTinterwordspacing
K.~Bestuzheva, M.~Besan{\c{c}}on, W.-K. Chen, A.~Chmiela, T.~Donkiewicz, J.~van
  Doornmalen, L.~Eifler, O.~Gaul, G.~Gamrath, A.~Gleixner, L.~Gottwald,
  C.~Graczyk, K.~Halbig, A.~Hoen, C.~Hojny, R.~van~der Hulst, T.~Koch,
  M.~L{\"u}bbecke, S.~J. Maher, F.~Matter, E.~M{\"u}hmer, B.~M{\"u}ller, M.~E.
  Pfetsch, D.~Rehfeldt, S.~Schlein, F.~Schl{\"o}sser, F.~Serrano, Y.~Shinano,
  B.~Sofranac, M.~Turner, S.~Vigerske, F.~Wegscheider, P.~Wellner, D.~Weninger,
  and J.~Witzig, ``{The SCIP Optimization Suite 8.0},'' Optimization Online,
  Technical Report, December 2021. [Online]. Available:
  \url{http://www.optimization-online.org/DB_HTML/2021/12/8728.html}
\BIBentrySTDinterwordspacing

\end{thebibliography}
\end{document}